\documentclass{spie}

\usepackage{cite}

\usepackage{times} 
\usepackage{indentfirst} 
\usepackage{soul}
\usepackage[colorinlistoftodos,color=pink]{todonotes} 
\usepackage{xkeyval}
\presetkeys{todonotes}{inline}{}
\usepackage{subcaption}

\usepackage{lipsum} 

\title{EANet: Enhanced Attribute-Based RGBT Tracker Network}

\author{Abbas Türkoğlu, Erdem Akagündüz
  \skiplinehalf
  Department of Modeling and Simulation, Graduate School of Informatics, METU, Ankara, Türkiye
    \skiplinehalf
    \normalsize
  \texttt{\{abbas.turkoglu,akaerdem\}@metu.edu.tr}
}

\begin{document}

\maketitle

\begin{abstract}
Tracking objects can be a difficult task in computer vision, especially when faced with challenges such as occlusion, changes in lighting, and motion blur. Recent advances in deep learning have shown promise in challenging these conditions. However, most deep learning-based object trackers only use visible band (RGB) images. Thermal infrared electromagnetic waves (TIR) can provide additional information about an object, including its temperature, when faced with challenging conditions. We propose a deep learning-based image tracking approach that fuses RGB and thermal images (RGBT). The proposed model consists of two main components: a feature extractor and a tracker. The feature extractor ecnodes deep features from both the RGB and the TIR images. The tracker then uses these features to track the object using an enhanced attribute-based architecture. We propose a fusion of attribute-specific feature selection with an aggregation module. The proposed methods is evaluated on the RGBT234 \cite{LiCLiang2018} and LasHeR \cite{LiLasher2021} datasets, which are the most widely used RGBT object-tracking datasets in the literature. The results show that the proposed system outperforms state-of-the-art RGBT object trackers on these datasets, with a relatively smaller number of parameters. 
  
  \keywords{Deep learning, RGBT Object Tracking, Computer Vision, Thermal and Visual Fusion}
\end{abstract}

\section{Introduction}

Tracking objects using both RGB and thermal images, known as RGBT tracking, is a complex task due to the differences in the two modalities. In challenging scenarios, the fusion of RGB and thermal images into a single stream using traditional tracking methods may not be the most effective approach. This work aims to enhance RGBT tracking through the use of an improved fusion module. With our approach, we aim to enhance feature fusion capability with a relatively small number of parameters and reduce dependency on large-scale training datasets.

For this purpose, we focus on developing attribute-specific fusion branches and attribute-based aggregation fusion modules, which are ideas previously proposed by \cite{APFNet}. In order to improve the effectiveness of this approach, we combine the attribute-specific fusion branch and the attribute-based aggregation fusion module with an architecture inspired by the so-called ESKNet \cite{Chen2022}. The aggregation module based on ESKNet is designed to fuse features from all branches adaptively. By adding branch features together, this module determines which features to add. It does this by using attention-based scoring to get rid of noisy features from attributes. The purpose of this paper is to combine this idea with attribute-specific fusion in order to improve RGBT tracking.

We evaluate the effectiveness of our work by comparing its performance against other state-of-the-art methods on benchmark datasets such as RGBT234\cite{LiCLiang2018}  and LasHeR \cite{LiLasher2021}. We aim to prove that our efforts provide a viable solution for enhancing the precision and speed of RGBT tracking systems across a diverse range of uses, such as autonomous driving, security surveillance, and robotics.

\section{Related Work}

When it comes to visual tracking, combining multiple imaging modalities has shown potential in overcoming the limitations of single-modal tracking systems. In recent years, RGBT tracking studies have increased significantly due to the availability of all-in-one RGB and TIR optical systems \cite{danaci2023survey}. In particular, deep learning-based RGBT architectures dominated the literature with the rise of deep learning. The reader may refer to \cite{Tang2022} for a detailed survey on the subject. In the following, we present details of several outstanding state-of-the-art RGBT object tracking methods that we used to benchmark our study. 

\begin{itemize}

\item\emph{MDNet}: An exemplary visual tracking algorithm is the Multi-Domain Network (MDNet) \cite{NamHan2015}. It delivers superior performance when compared to other algorithms in the field.  There are several branches of domain-specific layers in MDNet, and each domain corresponds to a separate training sequence. Each training video is treated as a separate domain, with domain-specific layers for binary classification at the end of the network. The algorithm also includes a multi-domain learning framework that separates domain-independent information from domain-specific information and an effective hard negative mining technique for the learning procedure.

\item\emph{FANet}: FANet \cite{ZhuY2018} offers a practical approach to implementing all layer features in a trained model and developing resilient target representations. It uses a fully connected layer to learn a nonlinear interaction between channels of varying modalities and employs a second fully connected layer in conjunction with a SoftMax activation function to predict the modality weights that regulate the information flows across modalities in adaptive aggregation.

\item\emph{DAPNet}: Dense feature Aggregation and Pruning network (DAPNet) \cite{DAPNet} offers a unique feature aggregation and pruning system for RGBT fusion object tracking, recursively combining all layers' deep features while compressing feature channels. 

 \item\emph{CMPP}: Cross-Modal Pattern-Propagation for RGBT Tracking (CMPP) \cite{Wang2020} diffuses instance patterns across the two modalities on both the spatial and temporal domain, incorporating long-term historical contexts into the current frame for more effective information inheritance. \cite{YangR2019} proposed a dual visual attention-guided deep RGBT tracking algorithm that utilizes both local attention and target-driven global attention.

\item\emph{JMMAC}: Joint Modeling of Motion and Appearance Cues (JMMAC) \cite{ZhangJ2021} is proposed for jointly modeling both appearance and motion cues. The framework includes multimodal fusion and motion mining components. When the appearance cue is unreliable, the motion cue, which includes target and camera movements, is used to make the tracker more robust.

\item\emph{MANet++}: MANet++ \cite{Lu2021} jointly performs modality-shared, modality-specific, and instance-aware target representation learning for RGBT tracking. Multimodal Cross-Layer Bilinear Pooling for RGBT Tracking (CBPNet) \cite{Xu2021} includes a feature extractor, channel attention mechanism, cross-layer bilinear pooling module, and three fully connected layers for binary classification. 
\end{itemize}

In addition to the listed above, M\begin{math}^5\end{math}L \cite{Tu2022}, MaCNet \cite{ZhangH2020}, TFNet \cite{zhu2021}, CAT \cite{Li2020}, ADRNet \cite{Zhang2021}, and APFNet \cite{APFNet} are various RGBT trackers that improve tracking performance by learning effective residual representations to enhance target appearance under various challenging circumstances. M\begin{math}^5\end{math}L \cite{Tu2022} uses a novel loss function called Multi-modal Multi-margin Structured Loss, which preserves structured information from both RGB and thermal modalities. MaCNet \cite{ZhangH2020} focuses on scene-adaptive fusion for cross-modal features, while TFNet \cite{zhu2021} tracks specific instances in consecutive frames using both RGB and thermal infrared information.

CAT \cite{Li2020}, ADRNet \cite{Zhang2021}, and APFNet \cite{APFNet} differ from most existing RGBT trackers in their network construction, making them more comprehensible for both modality-specific and modality-shared challenges. ADRNet \cite{Zhang2021} models target appearance in different attributes individually using the attribute-driven branch (ADRB), which is then adaptively aggregated via the channel-wise ensemble network (CENet) module. APFNet \cite{APFNet} is disentangling the fusion process into five challenge attributes, including thermal crossover, illumination variation, scale variation, occlusion, and fast motion. APFNet \cite{APFNet} is trained using a three-stage algorithm, using a dual-stream hierarchical architecture for online tracking and an enhancement transformer for interactive enhancement. These trackers are able to handle various challenges and can be trained efficiently using a small amount of data.

\section{Proposed Method}

\begin{figure}[t]
    \centering
    \includegraphics[width=\textwidth,height=\textheight,keepaspectratio]{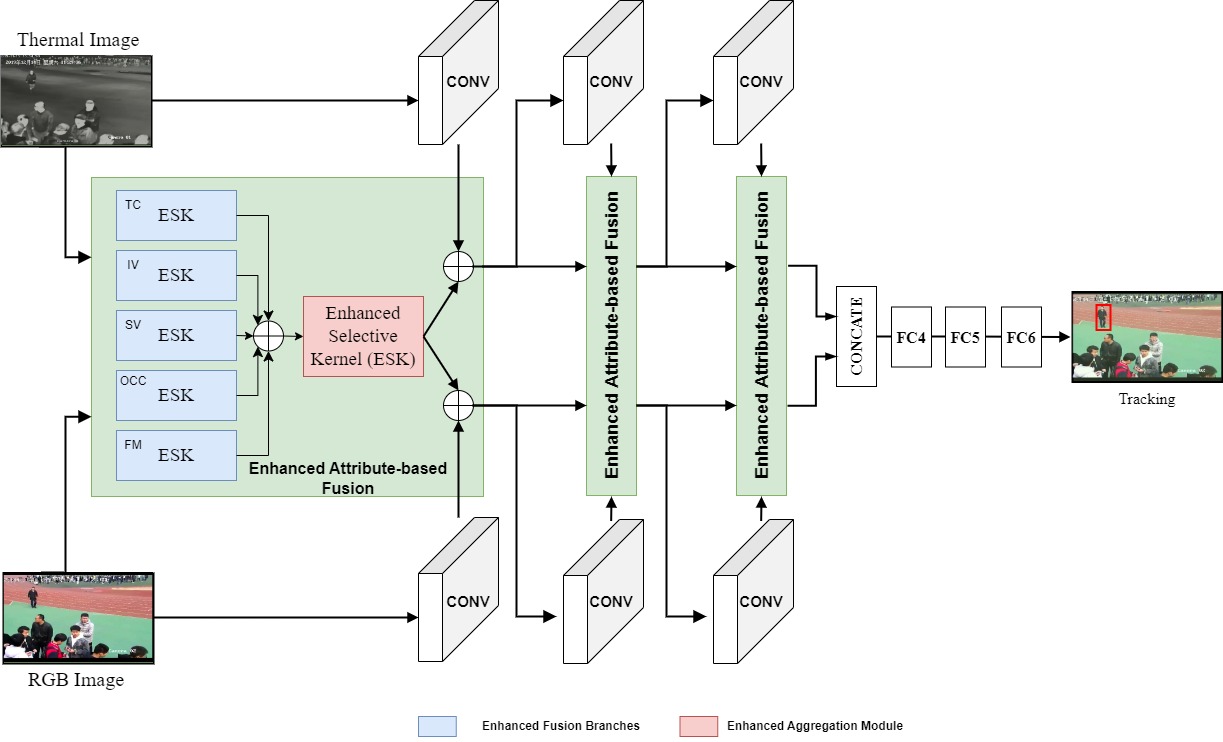}
    \caption{The structure of Enhanced Attribute-Based Network (EANet)}
    \label{fig:eanet}
\end{figure}

As stated earlier in the Introduction section, our research focuses on combining the attribute-specific fusion proposed by \cite{APFNet} with an aggregation module inspired by \cite{Chen2022}. The so-called ``Attribute-Based Progressive Fusion Module'' is utilized by APFNet for fusion. This module consists of three main components, namely: Attribute-Specific Fusion Branch, Attribute-based Aggregation, and Attribute-based Enhanced Fusion \cite{APFNet}. This architecture demonstrates to be a powerful RGBT tracker network, however, in this paper, we investigated ways to simplify the system complexity by replacing the Attribute-based Enhanced Fusion module.

Another architecture we are inspired by, the ESKNet, adds spatial attention mechanisms to a previous model, the SKNET \cite{SKNet}. Spatial attention enables the network to dynamically assign higher weights or focus to specific spatial locations, successfully calibrating the significance of various regions. The spatial attention mechanism prioritizes important regions for tracking by assigning them higher weights, while disregarding or suppressing regions that are less informative. The model can better capture the appearance of the object and adapt to changes in its position or appearance over time by focusing on the most important spatial locations. By calibrating the spatial dimension features through the integration of spatial attention, we hope to improve the model's ability to pay attention to and track the object correctly, even in tough situations like occlusions, cluttered backgrounds, or changing lighting conditions.

In order to accommodate for the differences between the various imaging modalities, we adopt a parallel network \cite{Li2018} as our network's backbone in order to extract features from both RGB and thermal infrared pictures independently. 
The backbone of our model is the first three convolution layers from VGG-M \cite{vgg-m}, which have kernel sizes of 7x7, 5x5, and 3x3. To initialize the convolution kernel parameters, we employ a pre-trained model on ImageNet-vid \cite{NamHan2015}. By incorporating the enhanced attribute-based module into all levels of the backbone, we create a hierarchical design that facilitates effective fusion of diverse modalities. Following the last convolutional layer, we introduce three fully connected layers, with the final FC6 layer adapted to different domains in a manner similar to the approach in \cite{NamHan2015}.

 The fusion process is applied to both RGB and thermal data simultaneously, passing through branches specially trained for different attributes. The Enhanced Selective Kernel (ESK) \cite{Chen2022} is used for both fusion branches and aggregation of branches data. The convolution data and the data formed as a result of the aggregation are subjected to element-wise addition, which proceeds separately for RGB and Thermal data in parallel.

\begin{figure}[t]
    \centering
    \begin{subfigure}{0.50\columnwidth}    \includegraphics[width=1\textwidth,height=\textheight,keepaspectratio]{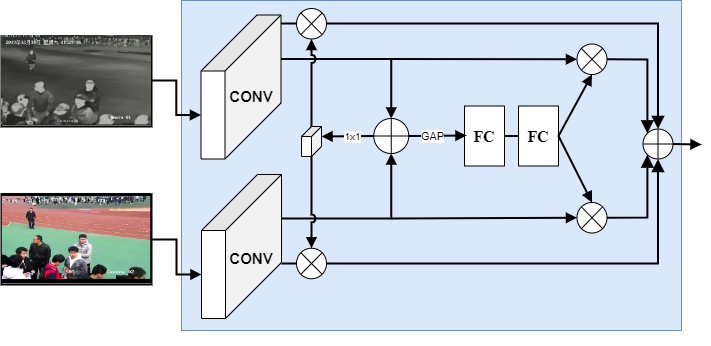}
        \caption{}
    \end{subfigure}
    \begin{subfigure}{0.49\columnwidth}	 \includegraphics[width=1\textwidth,height=\textheight,keepaspectratio]{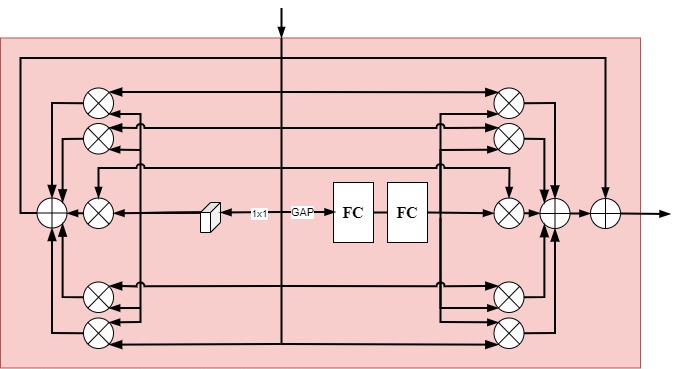}
        \caption{}
    \end{subfigure}
    \caption{ a) The structure of the proposed ``Enhanced Fusion Branch Module'' b) The structure of the proposed ``Enhanced Aggregation Module''}
    \label{fig:esk}
\end{figure}

In the interest of keeping things as straightforward as possible, we have mapped the fusion branches that fall under various attributes of the same structure. To be more specific, we begin by extracting features from two modalities for each branch by utilizing a convolutional layer with a kernel size of 5x5, a rectified linear unit (ReLU) \cite{Glorot2011}, and a convolutional layer with a kernel size of 4x4. This process is repeated for each branch. After that, we make use of a similar structure to ESK \cite{Chen2022} to make an adaptive selection of the features. Figure \ref{fig:eanet} and \ref{fig:esk} provide further information regarding these particulars.



The network is trained in two phases: first, each branch is trained individually, with the pre-trained model on the ImageNet dataset initializing the parameters of the VGG-M \cite{vgg-m} backbone. During the second phase, we rectify the trained branches and employ all the accessible training data to refine the aggregation fusion modules. This process involves a total of 500 training epochs. All the other settings remain the same as in the first phase. We store the parameters of the FC4 and FC5 aggregation fusion modules.

In the online tracking phase, the tracker is established using the target's location and the first frame of the series. 500 positive samples of various scales surrounding the target are collected using Gaussian sampling in the first frame, while 1000 samples are collected simultaneously to train the regressor. The coordinates of the tracking results are modified using the regressor to get more precise tracking results in follow-up tracking.

During the online tracking phase, only the parameters of the fully-connected layers are changed. Gaussian sampling and the tracking result from the \begin{math}t-1\end{math}-th frame acquire 256 samples for the current frame while tracking the target in the \begin{math}t\end{math}-th frame. The trained model computes the scores for 256 samples, then computes the mean value for the 5 samples with the highest scores at the time. Finally, the learned regressor fine-tunes the target location.

Long-time update Settings and short-time updates for tracking errors guarantee the robustness of the method. The formula for the tracking result is as follows: 

\begin{equation}
    X_t^*=\arg\max f^+(x_t^i)\:i=1,2,...,N
\end{equation}

\section{Experimental Setup}
 The codes for our model are implemented in PyTorch.
 The experiments were carried out on a system using an NVIDIA A4000 GPU with 16 GB RAM.
 \subsection{Datasets}
Our method underwent an extensive evaluation on the RGBT234 \cite{LiCLiang2018} dataset, which is widely recognized as a benchmark for RGB and thermal object tracking. The dataset comprises 234 high-resolution RGB and thermal image sequences, encompassing a diverse range of scenarios and challenges. The purpose of evaluating our method on the RGBT234 \cite{LiCLiang2018} dataset was to demonstrate its superior performance in terms of accuracy and effectiveness in tracking targets under challenging conditions when compared to several state-of-the-art RGBT trackers.

The other dataset we utilized in our experiments, namely LasHeR \cite{LiLasher2021} is comprised of 1224 (245 for testing and 979 for training) pairs of visible and thermal infrared video  and over 730K frame pairs in total. Each frame is aligned spatially and manually labeled with a bounding box, resulting in a heavily annotated dataset. It will play a vital role in both the training of deep RGBT fusion object trackers and the evaluation of RGBT fusion object tracking methods comprehensively.

The GTOT \cite{LiC2016} benchmark consists of 50 video sequences that were captured using both grayscale and thermal cameras. The sequences include a variety of challenging scenarios, such as low illumination, fast motion, partial occlusion, and cluttered backgrounds. The GTOT dataset also includes annotations of the object's bounding box in each frame, as well as statistics on the bias between the two modalities.


\begin{figure}[t]
    \centering
    \begin{subfigure}{0.49\linewidth}
\includegraphics[width=\textwidth,height=\textheight,keepaspectratio]{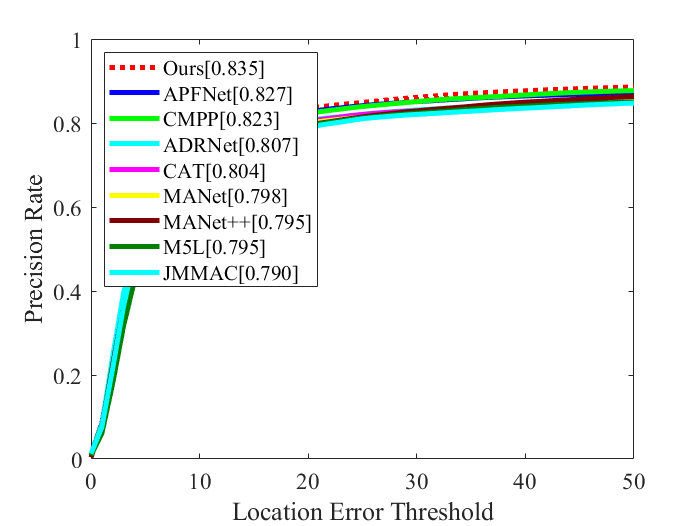}
        \caption{}
    \end{subfigure}
    \begin{subfigure}{0.49\linewidth}
\includegraphics[width=\textwidth,height=\textheight,keepaspectratio]{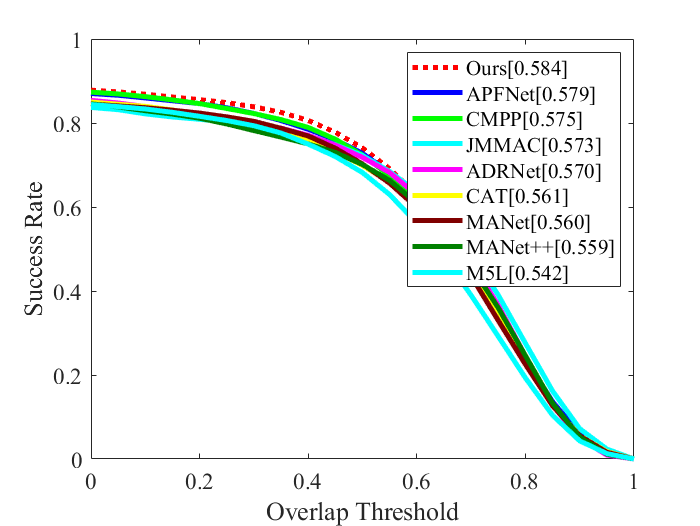}
        \caption{}
    \end{subfigure}
    \caption{The evaluation curve of Precision Rate (a) and Success Rate (b) on RGBT234 dataset.}
    \label{fig:rgbt234-PRSR}
\end{figure}

In this study, GTOT is used as the training set, and RGBT234  and LasHeR are used as the test sets.

\subsection{Evaluation Metrics}
Evaluation metrics for RGBT Tracking vary depending on the benchmark used. 
The GTOT \cite{LiC2016}, RGBT234 \cite{LiCLiang2018}, and LasHeR \cite{LiLasher2021} benchmarks all make use of the same metrics for evaluation, which are referred to as Precision Rate (PR) and Success Rate (SR). The distance between the actual bounding box on the ground and the one that was anticipated is what the precision rate measures. The term "success rate" refers to the proportion of failed tracking attempts whose Intersection over Union (IoU) values between their respective labels fall below a predetermined cutoff.  For quantitative performance evaluation, we use the PR and SR from the one-pass evaluation (OPE). In order to calculate the PR score, we adjust the threshold to 20 pixels. The area under the curve is used to produce the representative SR score, which expresses how many correctly tracked frames have overlaps that are greater than thresholds. 

\section{Results}

\subsection{Evaluation on RGBT234 Dataset}
In terms of overall performance, our method achieved remarkable results on the RGBT234 dataset. The precision (PR) score reached an impressive 83.5\%, showcasing the high level of accuracy in target tracking. Additionally, the success rate (SR) achieved an exceptional score of 58.4\%, indicating the effectiveness of our method in successfully tracking targets throughout the sequences. We compared our model with state-of-art models such as \cite{APFNet,Wang2020,Zhang2021,Li2020,LiCLu2019,Lu2021,Tu2022,ZhangJ2021}. The evaluation curve of Precision Rate and Success Rate can be seen in Figure \ref{fig:rgbt234-PRSR}.

To provide a comprehensive analysis of our method's performance, we further evaluated its effectiveness based on different attributes of the dataset. This attribute-based analysis allowed us to analyze how well our method handled specific challenges commonly encountered in object tracking, such as background clutter, occlusion, motion blur, and more.

Our method demonstrated robust performance across most attributes, surpassing the comparison trackers in terms of PR and SR. We compared our attribute-based performance with state-of-art treackers. Our attribute-based performance results can be seen in Table \ref{Tab:attribute-based-results}.

According to attribute-based performance results, in sequences with high background clutter (BC), our method achieved the best result with a precision rate of 83.2\% and a success rate of 54.6\%. This indicates its effectiveness in accurately tracking targets in cluttered environments. Our method performed well even in scenarios with heavy occlusion (HO), achieving a precision rate of 76.0\% and a success rate of 52.2\%. These results indicate that our approach is capable of dealing with challenging situations where the target is significantly occluded. Under motion blur (MB) conditions, our method achieved a precision rate of 76.9\% and a success rate of 55.7\%. This showcases its effectiveness in accurately tracking targets despite the presence of motion blur artifacts. When faced with partial occlusion (PO), our method achieved a precision rate of 86.6\% and a success rate of 60.6\%. This demonstrates the ability of our method to handle partially occluded targets and maintain accurate tracking. 
In sequences with thermal crossover (TC), our method achieved a precision rate of 82.8\% and the best success rate with a 59.2\% score. This showcases the effectiveness of our approach in accurately tracking targets when the thermal and RGB modalities overlap. 
Our method also demonstrated good performance in sequences with scale variation (SV), achieving second-best scores with a precision rate of 83.1\% and a success rate of 58.5\%. This highlights its ability to handle changes in target size and maintain accurate tracking. 

The attribute-based performance results provide comprehensive insights into the capabilities of our method in handling diverse challenges within the RGBT234 dataset. Our model gains almost 5 best scores over 12 attributes and the best score in all attributes. Our method outperformed the comparison trackers in terms of PR and SR and demonstrated its effectiveness in various challenging scenarios.

\begin{table}[t]
\fontsize{7.5pt}{7.5pt}\selectfont
\centering
\caption{Attribute-based PR and SR scores of RGBT234 dataset.  The best scores are marked in \textcolor{red}{red}.}{\label{Tab:attribute-based-results}}
\bigskip
\begin{tabular}{|p{\textwidth/25}|p{\textwidth/14}|p{\textwidth/14}|p{\textwidth/14}|p{\textwidth/14}|p{\textwidth/14}|p{\textwidth/14}|p{\textwidth/14}|p{\textwidth/14}|p{\textwidth/14}|}

    & JMMAC                        & M5L         & MANet++                      & MANet       & CAT         & ADRNet      & CMPP                         & APFNet                       & Ours                         \\

BC  & 0.687/0.485                  & 0.750/0.477 & 0.763/0.492                  & 0.781/0.512 & 0.811/0.519 & 0.804/0.536 & \textcolor{red}{0.832}/0.538 & 0.813/0.545                  & \textcolor{red}{0.832/0.546}                  \\
CM  & 0.762/0.556                  & 0.752/0.529 & 0.741/0.520                  & 0.726/0.518 & 0.752/0.527 & 0.743/0.529 & 0.756/0.541                  & \textcolor{red}{0.779/0.563}                  & 0.770/0.549                  \\
DEF & 0.706/0.529                  & 0.736/0.511 & \textcolor{red}{0.787}/0.562 & 0.739/0.530 & 0.762/0.541 & 0.743/0.528 & 0.750/0.541                  & 0.785/\textcolor{red}{0.564} & 0.779/0.554                  \\
FM  & 0.610/0.417                  & 0.728/0.465 & 0.694/0.455                  & 0.720/0.456 & 0.731/0.470 & 0.749/0.489 & 0.786/0.508                  & \textcolor{red}{0.791/0.511}                  & 0.763/0.492                  \\
HO  & 0.677/0.483                  & 0.665/0.450 & 0.711/0.482                  & 0.693/0.478 & 0.700/0.480 & 0.714/0.496 & 0.732/0.503                  & 0.738/0.507                  & \textcolor{red}{0.760/0.522}                  \\
LI  & 0.840/\textcolor{red}{0.588} & 0.821/0.547 & 0.796/0.543                  & 0.819/0.552 & 0.810/0.547 & 0.811/0.560 & \textcolor{red}{0.862}/0.584 & 0.843/0.569                  & 0.841/0.566                  \\
LR  & 0.771/0.517                  & 0.823/0.535 & 0.789/0.520                  & 0.812/0.543 & 0.820/0.539 & 0.838/0.562 & \textcolor{red}{0.865/0.571}                  & 0.844/0.565                  & 0.860/0.562                  \\
MB  & 0.751/0.549                  & 0.738/0.528 & 0.733/0.518                  & 0.720/0.520 & 0.683/0.490 & 0.733/0.532 & 0.754/0.541                  & 0.745/0.545                  & \textcolor{red}{0.768/0.557}                  \\
NO  & 0.932/\textcolor{red}{0.694} & 0.931/0.646 & 0.902/0.664                  & 0.893/0.650 & 0.932/0.668 & 0.916/0.660 & \textcolor{red}{0.956}/0.678 & 0.948/0.680                  & 0.934/0.672                  \\
PO  & 0.841/\textcolor{red}{0.611} & 0.863/0.589 & 0.830/0.591                  & 0.862/0.602 & 0.851/0.593 & 0.851/0.603 & 0.855/0.601                  & 0.863/0.606                  & \textcolor{red}{0.866}/0.606 \\
SV  & \textcolor{red}{0.837/0.616}                  & 0.796/0.542 & 0.792/0.573                  & 0.801/0.568 & 0.797/0.566 & 0.786/0.562 & 0.815/0.572                  & 0.831/0.579                  & 0.831/0.585                  \\
TC  & 0.749/0.526                  & 0.821/0.564 & 0.772/0.558                  & 0.754/0.545 & 0.803/0.577 & 0.796/0.586 & \textcolor{red}{0.835}/0.583 & 0.822/0.581                  & 0.828/\textcolor{red}{0.592} \\

ALL & 0.790/0.573                  & 0.795/0.542 & 0.795/0.559                  & 0.798/0.560 & 0.804/0.561 & 0.807/0.570 & 0.823/0.575                  & 0.827/0.579                  & \textcolor{red}{0.835/0.584}                  
\end{tabular}

\end{table}
            
\subsection{Evaluation on LasHeR Dataset}
\begin{figure}[t]
    \centering
    \begin{subfigure}{0.49\linewidth}
\includegraphics[width=\textwidth,height=\textheight,keepaspectratio]{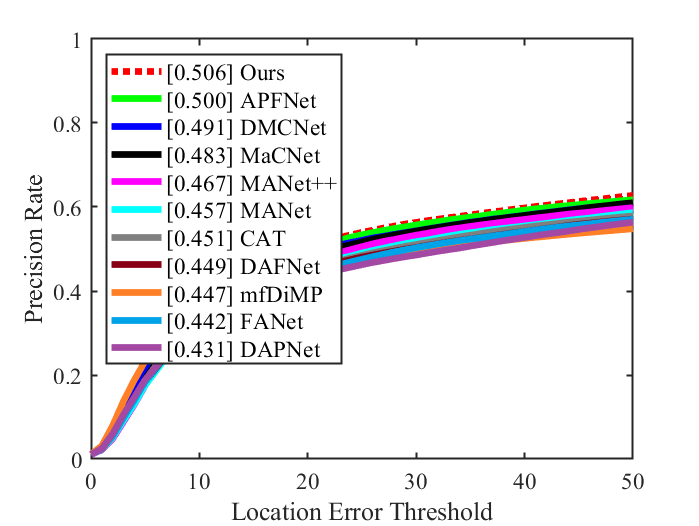}
        \caption{}
    \end{subfigure}
    \begin{subfigure}{0.49\linewidth}
\includegraphics[width=\textwidth,height=\textheight,keepaspectratio]{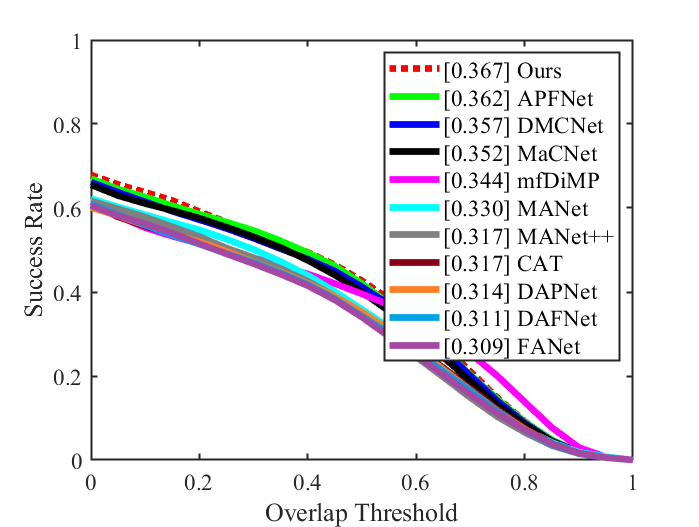}
        \caption{}
    \end{subfigure}
    \caption{The evaluation curves of Precision Rate (a) and Success Rate (b) on LasHer dataset.}
    \label{fig:Lasher-PRSR}
\end{figure}

In addition to the RGBT234 dataset, we also evaluated our method on the LasHeR dataset to further assess its performance.  The evaluation on this set was consistent, although with slightly lower performance compared to the RGBT234 dataset. The evaluation curve of Precision Rate and Success Rate on LasHeR dataset can be seen in Figure \ref{fig:Lasher-PRSR}. 
Our method achieved a PR of 50.6\% on the LasHeR dataset. Although this score is relatively lower compared to the RGBT234 dataset but also is the best score when compared to other trackers. It demonstrates the effectiveness of our approach in accurately localizing and tracking targets.
The SR achieved by our method on the LasHeR dataset is 36.7\%. While this score may seem lower, it reflects the challenges present in the LasHeR dataset and the ability of our approach to handle them to a significant extent. SR score of our model is also the best score when compared to other trackers.

The evaluation on the LasHeR dataset further supports the robustness and versatility of our method in handling different tracking scenarios and challenges. Despite the slightly lower performance compared to the RGBT234 dataset, our approach showcased its effectiveness in tracking targets in diverse environments.

\begin{table}[h]
 \centering
\caption{Results of Ablation Study on RGBT234 Dataset}
\begin{tabular}{|p{\textwidth/20}|p{\textwidth/8}|p{\textwidth/6}|}
\hline
   & Var-AggESK & Proposed Method  \\
\hline
PR &   \centering 0.812    &  0.835 \\
\hline
SR &  \centering 0.564    & 0.584 \\
\hline
\end{tabular}
\label{tbl:ablation}
\end{table}

\subsection{Ablation Study}

We conducted an ablation study to measure the effectiveness of the sub-modules of our model. In order to see the contribution of Enhanced Aggregation Module to the success of the model, we removed this module and aggregated the data from Enhanced Fusion Branch Modules with element-wise addition. We called this variation of the model ``Var-AggESK'' in Table \ref{tbl:ablation}. ``Var-AggESK'' variation achieved PR 81.2\% and SR 56.4\%. As seen in Table \ref{tbl:ablation}, when we compare ``Var-AggESK'' with the proposed model, we conclude that the Enhanced Aggregation Module contributes to the success of the model and is a necessary part of our model.

\section{Conclusion}

In conclusion, our method demonstrated outstanding performance on both the RGBT234 \cite{LiCLiang2018} and LasHeR \cite{LiLasher2021} datasets, highlighting its effectiveness in object tracking tasks. The evaluation on the RGBT234 \cite{LiCLiang2018} dataset showcased the high accuracy and effectiveness of our approach, with impressive precision and success rate scores. The attribute-based analysis further demonstrated its robustness in handling various challenges, including background clutter, occlusion, motion blur, and more.

While the evaluation on the LasHeR \cite{LiLasher2021} dataset yielded slightly lower performance scores compared to the RGBT234 dataset, but our model gains the best score when compared to other trackers, and it still confirmed the capabilities of our method in accurately tracking targets in diverse scenarios. The precision and success rate scores obtained on the LasHeR \cite{LiLasher2021} dataset demonstrate the effectiveness of our approach, considering the challenges specific to that dataset.

Overall, our method showcases significant potential in the field of RGBT object tracking. Its performance on the RGBT234 \cite{LiCLiang2018} and LasHeR \cite{LiLasher2021} datasets provides evidence of its accuracy, versatility, and ability to handle challenging scenarios. The results of our evaluation contribute to the advancement of RGBT object-tracking techniques and serve as a foundation for future research and development in the field.


\acknowledgements
This work is partially supported by Middle East Technical University Scientific Research Projects Coordination Unit (METU-BAP), under the project title "Real-Time Visual Tracking System based on Deep Learning using Infrared and Visible Band Fusion" (Kızılötesi ve Görünür Bant Kaynaştırma Kullanarak Derin Öğrenme Tabanlı ve Gerçek Zamanlı Görsel Takip Sistemi - AGEP-704-2022-11000).

\bibliographystyle{spiebib}
\bibliography{bibliography}



\section*{AUTHORS' BACKGROUND}

\begin{table}[ht]
	\centering
	\begin{tabular}{ | c | c | p{45mm} | c | } \hline
		Name & Title & \centering Research Field & Personal website \\
		\hline
		Abbas Türkoğlu & Master Student & Computer vision & \texttt{} \\
		\hline
		Erdem Akagündüz & Associate Professor & Deep Learning
Computer Vision & https://blog.metu.edu.tr/akaerdem\\
 		\hline
	\end{tabular}
\end{table}

\end{document}